%% file: Template.tex
\documentclass{article}
\usepackage{spconf,amsmath,graphicx}

\usepackage{amssymb}
\usepackage{booktabs}
\usepackage{float}
\usepackage{caption}
\usepackage{colortbl}
\usepackage{xcolor}
\usepackage{float}
\usepackage[export]{adjustbox}

\title{RayFormer: Modeling Inter- and Intra-Ray Similarity for NeRF-Based Video Snapshot  Compressive Imaging}
\name{Yubo Dong $^{1, 2, 3}$ \qquad Danhua Liu  $^{4,\star}$  \qquad Anqi Li $^4$ \qquad Zhenyuan Lin $^{4}$
\thanks{$\star$: Corresponding author (dhliu@xidian.edu.cn)}
}

\address{$^1$ Changzhi Medical College, Changzhi 046000, China \vspace{0.5em} \\
$^2$ Engineering Research Centre for Intelligent Data Assisted Diagnosis and Treatment \\
in Shanxi Province, Changzhi 046000, China \vspace{0.5em}\\
$^3$ Uniwave Artificial Intelligence Technology Co., Ltd., Changzhi 046000, China \vspace{0.5em} \\
$^4$ Xidian University, School of Artificial Intelligence, Xi’an 710126, China}

\begin{document}
%
\maketitle
\begin{abstract}
Video snapshot compressive imaging (SCI) enables the reconstruction of dynamic scenes from a single snapshot measurement. Recently, NeRF-based methods have shown promising reconstruction performance. However, such methods typically adopt random ray sampling strategies and fail to capture content structural similarities, resulting in limited reconstruction quality. To address these issues, we \textbf{first} propose a patch-level ray sampling strategy to enable the modeling of content structure. \textbf{Then}, we propose an Inter- and Intra-Ray Transformer (RayFormer) to capture the structural similarities—modeling both inter-ray similarities among spatially neighboring points at the same depth and intra-ray correlations between adjacent points along the viewing ray. \textbf{Finally}, benefiting from the patch-level sampling strategy, the total variation prior is incorporated into the objective function to enhance spatial smoothness and suppress artifacts. Experiments in both simulated and real-world scenes demonstrate that the proposed method achieves state-of-the-art (SOTA) reconstruction performance.
\end{abstract}


\vspace{-2mm}
\section{Introduction}
\vspace{-2mm}

Video Snapshot Compressive Imaging (SCI) \cite{cacti, yuan2021snapshot} has emerged as a promising computational imaging paradigm that enables the acquisition of high-speed video through a single 2D measurement. By encoding temporal information into spatially multiplexed patterns via designed coded masks—such as modulated patterns across time—video SCI effectively compresses multiple video frames into a single snapshot. During reconstruction, advanced algorithms are employed to disentangle and recover the original high-speed frames from the compressed measurement.

In recent years, video SCI has seen significant progress with various reconstruction methods, including model-based optimization techniques \cite{gap-tv, desci}, Plug-and-Play (PnP) frameworks \cite{pnp-ffdnet}, and end-to-end networks \cite{efficientsci, stformer}. Yet, these methods typically operate on 2D frames or 3D tensors without explicitly reconstructing the underlying 3D scene structure, limiting their ability to ensure geometric and structural consistency across views and time. Recently, the Neural Radiance Field (NeRF) has been adapted to video SCI reconstruction \cite{scinerf}, offering a continuous and view-coherent scene representation by modeling the 4D spatio-temporal radiance field. However, despite their representational power, current NeRF-based methods still rely on random or uniform ray sampling strategies that ignore content structural similarities in natural scenes. Moreover, they lack dedicated mechanisms to capture both inter-ray similarities among neighboring spatial locations and intra-ray correlations along the ray path—both of which are crucial for accurate and coherent reconstruction.

To address these limitations, we propose a structured and geometry-aware framework for NeRF-based video SCI reconstruction, which fundamentally rethinks how rays are sampled and how structural dependencies are modeled during the rendering process. First, we introduce a patch-level ray sampling strategy that replaces the conventional random or uniformly random sampling with a spatially coherent selection mechanism. Instead of treating each ray independently, our approach samples rays in structured patches—mimicking the local continuity of natural scenes—and ensures that neighboring rays are jointly considered during training and inference. This design explicitly preserves local spatial and temporal structural similarities, enabling more efficient scene exploration and promoting geometric consistency across adjacent regions. Designed to capture structural coherence, our patch-level ray sampling strategy ensures that structurally related regions are jointly sampled, thereby mitigating artifacts arising from sparse and unstructured ray selection. 

\begin{figure*}[!htb]
    \centering
    \includegraphics[width=\linewidth]{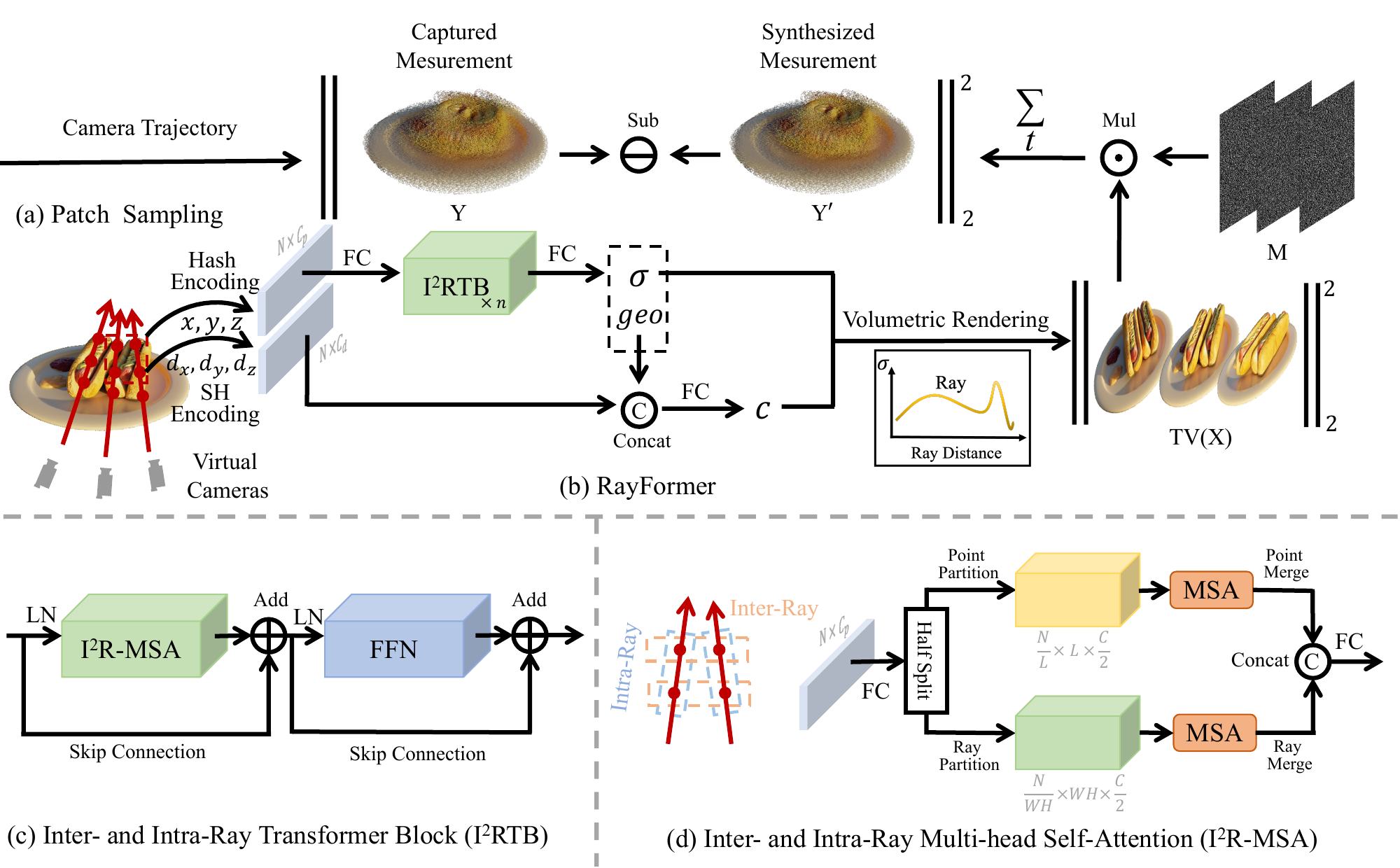}
    \vspace{-7mm}
    \caption{\small The overall architecture of the proposed method is illustrated, including patch-level ray sampling, RayFormer, I²RTB, and I²R-MSA.} 
    \vspace{-5mm}
    \label{fig: RayFormer}
\end{figure*}

Building upon this structured sampling scheme, we further propose the Inter-and Intra-Ray Transformer (RayFormer), a novel attention-based architecture specifically designed to capture both inter-ray and intra-ray structural dependencies within the 4D spatio-temporal radiance field. Unlike standard Transformers \cite{transformer, rdluf, dgsmp, adrnn} that operate on flattened sequences of points, RayFormer explicitly disentangles two distinct types of correlations: (i) inter-ray similarities, which model the contextual relationships among spatially neighboring points at the same depth across multiple rays, thereby enforcing spatial smoothness and structural coherence; and (ii) intra-ray correlations, which capture the sequential dependencies between adjacent 3D points along the viewing ray, reflecting the physical continuity of light transport and occlusion boundaries. By integrating dual-path attention mechanisms to jointly learn these two complementary priors, RayFormer enables more accurate density and color estimation.

Finally, the patch sampling strategy enables the effective incorporation of a total variation (TV) prior into the objective function, which promotes spatial smoothness and suppresses artifacts. This regularization complements the structural modeling capabilities of our network, leading to more faithful and visually plausible reconstructions. 

Overall, our contributions can be summarized as:
\begin{itemize}
    \item We propose a patch-level ray sampling strategy to enable the modeling of content structural similarities. 
    \item An Inter- and Intra-Ray Transformer (RayFormer) is proposed to capture structural similarities—modeling both inter-ray similarities among spatially neighboring points at the same depth and intra-ray correlations between adjacent points along the viewing ray.
    \item Benefiting from the patch-level sampling strategy, we incorporate the TV prior into the objective function to enhance spatial smoothness and reduce artifacts.
\end{itemize}

\section{Methodology}
\label{sec:format}

\subsection{Preliminaries}

\subsubsection{Imaging Model of Video SCI}

In video SCI, the high-dimensional spatio-temporal information of a scene is compressed into a single two-dimensional measurement. This acquisition process is physically realized by employing programmable optical devices—most commonly Digital Micromirror Devices (DMD) or liquid crystal-based spatial light modulators (SLM)—which rapidly encode the incident light during a single sensor exposure. Mathematically, the captured measurement $\mathbf{Y} \in \mathbb{R}^{H \times W}$ is formed by modulating $N$ consecutive latent frames $\{\mathbf{X}_i\}_{i=1}^N$ with a sequence of high-speed modulation masks $\{\mathbf{M}_i\}_{i=1}^N$ to $N$ within a single exposure sensor integration. This forward imaging process is defined as:
\begin{equation}
    \mathbf{Y} = \sum_{i=1}^{N} \mathbf{X}_i \odot \mathbf{M}_i + \mathbf{Z},
    \label{eq: video sci}
\end{equation}
where $H$ and $W$ are the image's spatial dimensions, $\odot$ denotes Hadamard (element-wise) product, and $\mathbf{Z}$ accounts for the additive measurement noise. The temporal compression ratio $N$ is determined by the number of unique modulation patterns projected during the sensor's exposure time.

\subsubsection{NeRF-based Video SCI Reconstruction}

NeRF-based video SCI framework \cite{scinerf} jointly optimizes 3D scene representations and camera poses \cite{barf, bad_nerf, nerf--} from a single compressed measurement. Poses $\mathbf{T}_i$ are initialized near identity with small perturbations and modeled via SE(3) interpolation:   
\begin{equation}
    \mathbf{T}_i = \mathbf{T}_1 \exp\left(\frac{i}{N} \log(\mathbf{T}_1^{-1} \mathbf{T}_N)\right),
    \label{eq: virtual camera pose}
\end{equation}

NeRF predicts volume density $\sigma(\mathbf{r}(t))$ and view-dependent color $\mathbf{c}(\mathbf{r}(t),\mathbf{d})$ from 5D point coordinates, and renders pixel values via volumetric integration:  
\begin{equation}
    \begin{aligned}
        & C(\mathbf{r}) = \int_{t_n}^{t_f} T(t) \sigma(\mathbf{r}(t)) \mathbf{c}(\mathbf{r}(t), \mathbf{d}) \, dt, \\
        & T(t) = \exp\left(-\int_{t_n}^t \sigma(\mathbf{r}(s))\,ds\right).
    \end{aligned}
    \label{eq: volumetric rendering}
\end{equation}

The loss minimizes the difference between captured and reconstructed measurements:  
\begin{equation}
    \mathcal{L} = \sum_{\mathbf{r} \in \mathcal{R}} \left\| \mathbf{Y}(\mathbf{r}) - \sum_{i=1}^{N} \mathbf{M}(\mathbf{r}, i) \odot C(\mathbf{r}, i) \right\|^2,
    \label{eq: loss}
\end{equation}
where $\mathcal{R}$ is the ray set, and $C(\mathbf{r}, i)$ is the $i$-th frame’s rendering under mask $\mathbf{M}(\mathbf{r}, i)$.

\subsection{RayFormer}

The overall architecture is shown in Fig.~\ref{fig: RayFormer}. First, we propose patch-level ray sampling to enable content structural similarities modeling. Then, point and direction features $P \in \mathbb{R}^{N \times C_p}$ and $D \in \mathbb{R}^{N \times C_d}$ are obtained via hash grid and spherical harmonics encoding. Subsequently, RayFormer processes $P$ through Inter- and Intra-Ray Attention Blocks (I²RAB) to predict density $\sigma$ and geometry features $geo$, with the latter concatenated with $D$ and fed to an FC layer for predicting color $C$. Ultimately, volumetric rendering uses density $\sigma$ and color $C$ to synthesize virtual frames $X$, which are then modulated by masks $M$ to generate the measurement $Y'$. Finally, the loss $\|Y' - Y\|^2$ is augmented with a total variation loss enabled by patch-level ray sampling.

\subsubsection{Patch-level Ray Sampling}

In NeRF-based video SCI, current methods rely on random or uniform ray sampling strategies. While effective for simple geometries, this "point-wise" approach inherently neglects the intrinsic spatial coherence and structural continuity essential to real-world scenes. By treating pixels as independent entities, random sampling fails to capture vital local geometric priors and high-frequency textures. Without such spatial awareness, standard SCINeRF models often produce reconstructions that lack fine-grained sharpness and suffer from fragmented spatial artifacts.

To overcome these limitations and better leverage the spatio-temporal correlations inherent in video SCI, we propose a patch-level ray sampling strategy (Fig.~\ref{fig: RayFormer}(a)). Unlike traditional methods that sample isolated rays, our approach partitions the radiance field into overlapping windows that preserve neighborhood relationships and local context. During the training phase, these windows are sampled stochastically for optimization, forcing the model to learn from dense, localized structures rather than disjointed points. At the inference stage, the framework employs window-wise processing to reconstruct high-fidelity virtual frames across all poses. This methodology not only ensures global consistency but also significantly enhances the recovery of fine-grained spatial details.


\input{Tables/table1}
\begin{figure*}[h]
    \centering
    \includegraphics[width=\linewidth]{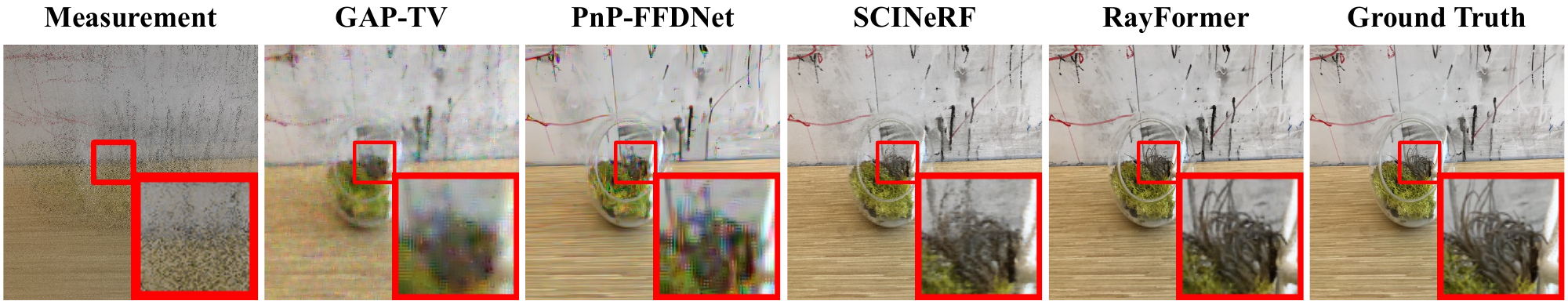}
    \caption{\small Visual comparison of RayFormer with SOTA video SCI reconstruction methods on the ``airplants" scene.} 
    \vspace{-3mm}
    \label{fig: simu_results}
\end{figure*}

\subsubsection{Inter- and Intra-Ray Multi-head Self-Attention}


The Intra- and Inter-Ray Multi-Head Self-Attention (I$^2$R-MSA) module serves as the core component of RayFormer. Its primary objective is to transcend standard point-wise processing by simultaneously capturing inter-ray similarities (spatial context across different rays) and intra-ray correlations (depth-wise relationships along a ray).

Initially, the input point features $X \in \mathbb{R}^{N \times C}$ are transformed into query ($Q$), key ($K$), value ($V$) embeddings via three independent fully connected (FC) layers, such that $Q, K, V \in \mathbb{R}^{N \times C}$. Then, as illustrated in Fig.~\ref{fig: RayFormer}(d), the query $Q$, key $K$, and value $V$ are half split along the channel dimension into two equal parts:  
\begin{equation}
    \begin{aligned}
        & Q, K, V \in \mathbb{R}^{N \times C} \mapsto \\
    & (Q_{a}, Q_{b}), (K_{a}, K_{b}), (V_{a}, V_{b}) \in \mathbb{R}^{N \times \frac{C}{2}} \times \mathbb{R}^{N \times \frac{C}{2}}.
    \end{aligned}
    \label{eq: halfsplit}
\end{equation}
To facilitate specialized attention mechanisms, these split features are reshaped into distinct geometric tensors tailored for intra- and inter-ray modeling:
\begin{itemize}
    \item \textbf{Intra-ray Path}: Point features are grouped along the depth dimension. Let $L$  denote the number of points per ray for intra-ray self-attention. The tensors are reshaped to $\mathbb{R}^{\frac{N}{L} \times L \times \frac{C}{2}}$, where self-attention is performed across the $L$ points to model occlusion and density gradients along a single ray.
    \item \textbf{Inter-ray Path}: Point features are grouped within local spatial neighborhoods. Given a spatial window of size $W \times H$, the tensors are reshaped to $\mathbb{R}^{\frac{N}{WH} \times WH \times \frac{C}{2}}$. This allows the model to aggregate contextual information across $WH$ adjacent rays at corresponding depth levels.
\end{itemize}
The formal transformation is defined as:
\begin{equation}
    \begin{aligned}
    Q_{\text{intra}}, K_{\text{intra}}, V_{\text{intra}} &= \text{Reshape}(Q_a, K_a, V_a) \in \mathbb{R}^{\frac{N}{L} \times L \times \frac{C}{2}}, \\
    Q_{\text{inter}}, K_{\text{inter}}, V_{\text{inter}} &= \text{Reshape}(Q_b, K_b, V_b) \in \mathbb{R}^{\frac{N}{WH} \times WH \times \frac{C}{2}},
    \end{aligned}
\end{equation}
Following the partition, we apply standard Multi-Head Self-Attention (MSA) to both paths.
\begin{itemize}
    \item \textbf{Intra-ray attention} focuses on relationships among points along the same ray, effectively learning the "importance" of different depth intervals for color integration.
    \item \textbf{Inter-ray attention} captures correlations among rays within a spatial window at the same depth, allowing the model to "borrow" features from neighboring rays to ensure spatial structural consistency.
\end{itemize}
By jointly modeling these interactions, RayFormer effectively captures complex 3D scene structures, enabling more accurate reconstruction of the implicit neural radiance field. 



\subsubsection{Total Variance Loss Funtion}

Benefiting from patch-level ray sampling, we introduce the total variation (TV) prior into the loss to explicitly regularizes spatial smoothness. Unlike random ray sampling, our window-based loss accounts for local structural coherence and is formulated as:
\begin{equation}
    \mathcal{L} = \sum_{\mathbf{r_w} \in \mathcal{R}} \Bigg\| \mathbf{Y}(r_w) - \hat{\mathbf{Y}}(r_w) \Bigg\|_2^2 + \lambda \mathbf{TV}(C(r_w))
    \label{eq: loss_tv}
\end{equation}
where $\mathbf{r}_w$ denotes a ray patch, $\mathbf{Y}$ the corresponding measurement, and $\lambda$ balances data fidelity and regularization.

\section{Experiments}

\vspace{-2mm}
\subsection{Experimental Settings}

\textbf{Datasets}. Following \cite{scinerf}, we evaluate on six synthetic scenes: Airplants \cite{llff}, Hotdog \cite{nerf360}, Cozy2room, Tanabata, Factory, and Vendor \cite{deblurnerf}. To assess generalization, we further test on real-world SCI data captured by the setup in \cite{scinerf}.

\textbf{Compared methods and evaluation metrics}. We compare with several SOTA SCI reconstruction methods: GAP-TV (model-based) \cite{gap-tv}, PnP-FFDNet (plug-and-play) \cite{pnp-ffdnet}, EfficientSCI (end-to-end network) \cite{efficientsci}, and SCINeRF (NeRF-based) \cite{scinerf}. Performance is evaluated with the fidelity metric PSNR, the structural metric SSIM, and the perceptual metric LPIPS. 

\vspace{-2mm}
\subsection{Results}

\textbf{Performance on synthetic dataset.} Quantitative results in Table \ref{tab: simu_results} show that RayFormer overall outperforms competing methods, achieving the best PSNR on five of six scenes, with improvements over SCINeRF of up to 0.65 dB, 0.80 dB, 2.56 dB, and 4.22 dB on Vender, Airplants, Hotdog, and Tanabata, respectively. Qualitative results in Fig. \ref{fig: simu_results} show that GAP-TV suffers from severe noise and structural loss; PnP-FFDNet reduces noise but introduces artifacts and color distortions; SCINeRF recovers coarse shapes but yields blurry and inconsistent details. In contrast, our method reconstructs sharper edges and more faithful textures.

\begin{figure}[!t]
    \centering
    \includegraphics[width=\linewidth]{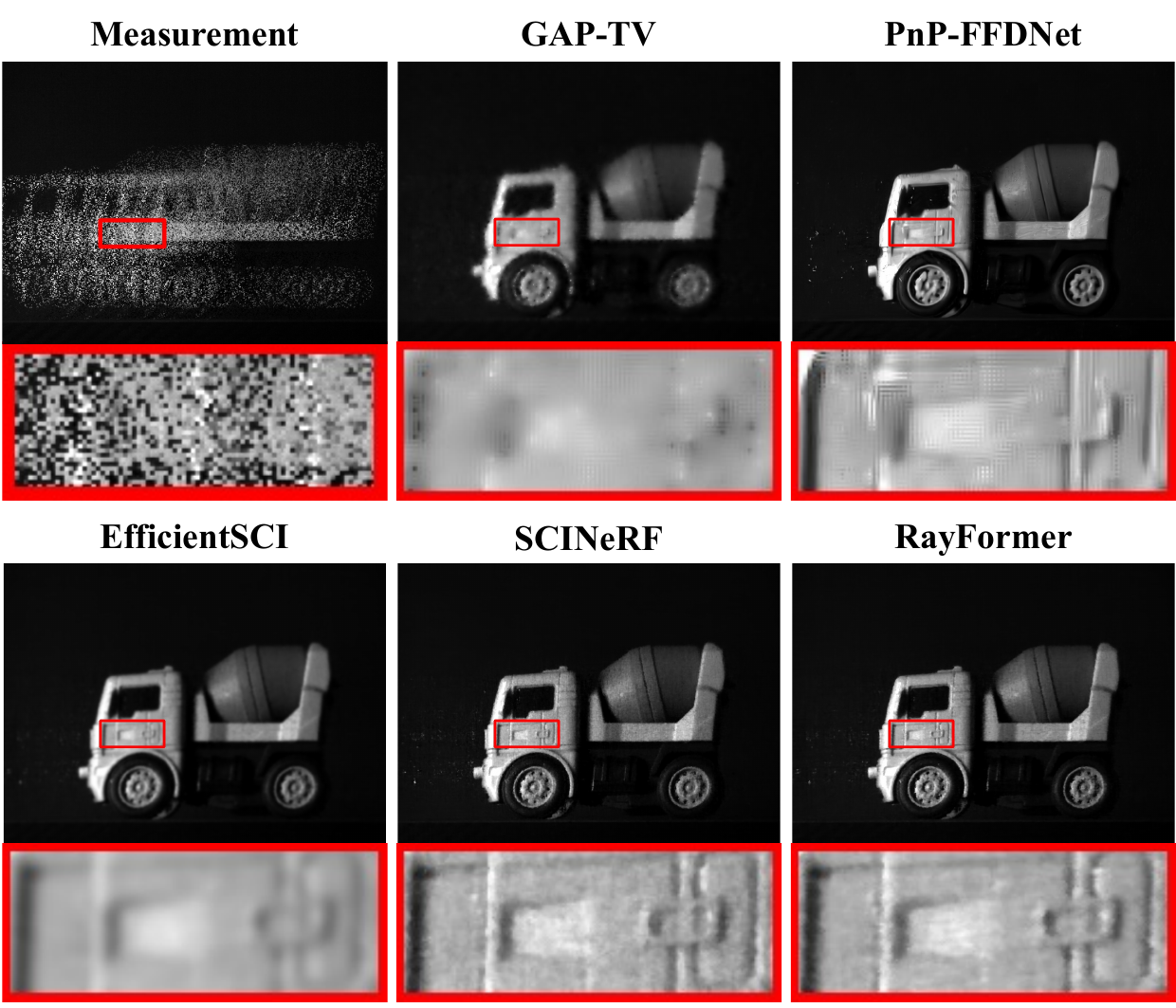}
    \caption{\small RayFormer vs. SOTA on the real ``truck" scene.} 
    \vspace{-3mm}
    \label{fig: real_results}
\end{figure}

\textbf{Performance on real-world data.} As shown in Fig. \ref{fig: real_results}, GAP-TV suffers from over-smoothing, making details unrecognizable. PnP-FFDNet introduces blurring and artifacts, while EfficientSCI produces smooth but blurry textures. SCINeRF improves edge clarity but retains noticeable noise. In contrast, our method recovers sharp structures and clear textures with minimal noise, demonstrating superior capability in recovering structural contents.

\subsection{Ablation Study}

\textbf{Effectiveness of Architectural Components}. To quantify the individual contribution of each proposed module, we conduct an incremental ablation study, as summarized in Table~\ref{tab: ablation_study}. Starting from a baseline PSNR of 30.11 dB, the introduction of \textbf{patch-level sampling} yields a 0.77 dB improvement, which we attribute to the structured distribution of sampled points that provides richer local context during training. The integration of \textbf{inter-ray attention} results in a substantial gain of 2.36 dB, underscoring its pivotal role in capturing structural similarities across different content regions. Furthermore, \textbf{intra-ray attention} contributes an additional 1.07 dB, validating its efficacy in modeling occlusion-aware coherence along individual rays. Finally, incorporating \textbf{TV loss} provides a marginal yet important boost of 0.73 dB by imposing spatial smoothness priors, which effectively suppresses visual artifacts. These results collectively demonstrate that each component is essential for the superior performance of RayFormer.

\input{Tables/ablation_study}

\textbf{Impact of Patch Size}. We further investigate the trade-off between reconstruction quality and computational efficiency regarding patch size. In our implementation, we employ overlapping patches during training and non-overlapping patches during inference to accelerate rendering. As shown in Table~\ref{tab: ablation1}, an $8 \times 8$ patch size achieves the optimal balance; smaller patches fail to capture sufficient local structure, while larger patches increase memory overhead without yielding significant performance gains.

\input{Tables/ablation1}

\textbf{Efficiency and Comparison with SCINeRF}. Beyond synthesis quality, we evaluate the computational efficiency of RayFormer against the state-of-the-art SCINeRF. Table~\ref{tab: ablation2} reports a comprehensive comparison across parameters, GFlops, memory usage, and inference latency. Thanks to its streamlined architecture—comprising only two transformer blocks with a hidden dimension of 32—and the elimination of the computationally expensive importance sampling stage, RayFormer achieves a significantly more lightweight footprint and faster inference speed while maintaining competitive accuracy.

\input{Tables/ablation2}

\section{Conclusion}

In this paper, we proposed patch-level ray sampling and the Inter- and Intra-Ray Transformer (RayFormer) to capture content structural similarities for NeRF-based Video SCI. Additionally, benefiting from the patch-level sampling strategy, we incorporated the total variation prior into the objective function to enhance spatial smoothness and reduce artifacts. Experiments on both synthetic and real scenes demonstrated the effectiveness of the proposed method.

\vfill\pagebreak

\bibliographystyle{IEEEbib}
\bibliography{strings,refs}

\end{document}

%% file: Tables/table1.tex
\begin{table*}[h]
	\setlength\tabcolsep{2pt}
	\parbox{\textwidth}{
		\resizebox{\linewidth}{!}{
		\begin{tabular}{c|ccc|ccc|ccc|ccc|ccc|ccc}
			\specialrule{0.1em}{1pt}{1pt}
			& \multicolumn{3}{c|}{Airplants} & \multicolumn{3}{c|}{Hotdog} & \multicolumn{3}{c|}{Cozy2room} & \multicolumn{3}{c|}{Tanabata} & \multicolumn{3}{c|}{Factory} & \multicolumn{3}{c}{Vender} \\
			& PSNR$\uparrow$ & SSIM$\uparrow$ & LPIPS$\downarrow$ & PSNR$\uparrow$ & SSIM$\uparrow$ & LPIPS$\downarrow$ & PSNR$\uparrow$ & SSIM$\uparrow$ & LPIPS$\downarrow$ & PSNR$\uparrow$ & SSIM$\uparrow$ & LPIPS$\downarrow$ & PSNR$\uparrow$ & SSIM$\uparrow$ & LPIPS$\downarrow$ & PSNR$\uparrow$ & SSIM$\uparrow$ & LPIPS$\downarrow$ \\
			\specialrule{0.05em}{1pt}{1pt}
			GAP-TV \cite{gap-tv} &22.85 &.4057 &.4986 & 22.35 &.7663 &.3179 & 21.77 &.4321 &.6031 & 20.42 &.4264 &.6250 & 24.05 &.5666 &.5149 &20.00 &.3678 &.6882\\
			PnP-FFDNet \cite{pnp-ffdnet} &27.79 &.9117 &.1817 &29.00 &.9765 &.0511 &28.98 &.8916 &.0984 & 29.17 & .9032 & .1197 & 31.75 & .8977 & .1142 & 28.70 & .9235 & .1315 \\
			EfficientSCI \cite{efficientsci} &30.13 & \underline{.9425} & .1129 & 30.75 & .9568 & .0461 & 31.47 & .9327 & .0476 & 32.30 & .9587 & .0600 & 32.87 & .9259 & .0709 & 33.17 & .9401 & .0456 \\
                SCINeRF \cite{scinerf} & \underline{30.69} &.9335 & \underline{.0728} & \underline{31.35} & \textbf{.9878} & \underline{.0310} & \underline{33.23} & \underline{.9492} & \underline{.0445} & \underline{33.61} & \underline{.9638} & \underline{.0374} & \textbf{36.60} & \underline{.9638} & \textbf{.0221} & \underline{36.40} & \underline{.9840} & \underline{.0298} \\ 
			\specialrule{0.05em}{1pt}{1pt}
                RayFormer & \textbf{31.49} & \textbf{.9659} & \textbf{.0319} & \textbf{33.92} & \underline{.9788} & \textbf{.0167} & \textbf{33.37} & \textbf{.9747} & \textbf{.0202} & \textbf{37.83} & \textbf{.9859} & \textbf{.0145} & \underline{36.58} & \textbf{.9804} & \underline{.0305} & \textbf{37.05} & \textbf{.9851} & \textbf{.0190} \\
			\specialrule{0.1em}{1pt}{1pt}
	\end{tabular}}
    \vspace{-0.7em}
	\caption{Quantitative comparison of RayFormer with SOTA video SCI reconstruction methods on synthetic datasets}
	\label{tab: simu_results}}
\end{table*}

%% file: Tables/ablation_study.tex
\begin{table}[h]
    \begin{tabular}{c|c|c|c|c}
    \specialrule{0.1em}{1pt}{1pt}
      & Components          & PSNR  & SSIM  & LPIPS \\
    \hline
    1 & baseline            & 30.11 & .9405 & .0554 \\
    2 & 1 + patch sampling  & 30.88 \color{green}{(+0.77)} & .9464 & .0504 \\
    3 & 2 + inter-ray       & 33.24 \color{green}{(+2.36)} & .9658 & .0352 \\
    4 & 3 + intra-ray       & 34.31 \color{green}{(+1.07)} & .9704 & .0282 \\
    5 & 4 + tv loss         & 35.04 \color{green}{(+0.73)} & .9785 & .0221 \\
    \specialrule{0.1em}{1pt}{1pt}
    \end{tabular}
    \caption{Ablation study of RayFormer components.}
    \vspace{-3mm}
    \label{tab: ablation_study}
\end{table}


%% file: Tables/ablation1.tex
\begin{table}[h]
    \begin{tabular}{c|c|c|c|c}
    \specialrule{0.1em}{1pt}{1pt}
    Patch Size      & $2 \times 2$  &  $4 \times 4$  & $8 \times 8$ & $16 \times 16$ \\
    \hline
    PSNR            & 31.23       & 33.43        & 35.04      & 34.01       \\
    \specialrule{0.1em}{1pt}{1pt}
    \end{tabular}
    \caption{Ablation study on patch size.}
    \label{tab: ablation1}
\end{table} 


%% file: Tables/ablation2.tex
\begin{table}[h]
    \begin{tabular}{c|c|c|c|c}
    \specialrule{0.1em}{1pt}{1pt}
    Method        & Params  &  GFlops  & Memory      & Inference time \\
    \hline
    SCINeRF       & 5.3M    & 146.63   & 14.29G      & 0.115s       \\
    \hline
    RayFormer     & 1.9M    & 20.18    & 6.598G      & 0.043s       \\
    \specialrule{0.1em}{1pt}{1pt}
    \end{tabular}
    \caption{Efficiency analysis against SCINeRF.}
    \label{tab: ablation2} 
\end{table}